\documentclass[journal]{IEEEtran}
\usepackage{multirow}
\usepackage{amsfonts}

\usepackage{cite}

\ifCLASSINFOpdf
  \usepackage[pdftex]{graphicx}
\else
  \usepackage[dvips]{graphicx}
\fi
\usepackage{amsmath}
\usepackage{algorithm}
\usepackage{algorithmic}

\usepackage{cases}
\usepackage{bm}
\usepackage{multirow}
\usepackage{subfig}
\usepackage{tabularx}
\newcolumntype{Y}{>{\centering\arraybackslash}X}

\begin{document}
%
\title{Deep Reinforcement Learning for Image Hashing}
%
%
%

	\author{
	Yuxin Peng, Jian Zhang and Zhaoda Ye
	\thanks{This work was supported by National Natural Science Foundation of China under Grant 61771025 and Grant 61532005.}
	\thanks{The authors are with the Institute of Computer Science and Technology, Peking University, Beijing 100871, China. Corresponding author: Yuxin Peng (e-mail: pengyuxin@pku.edu.cn).}
}

\maketitle

\begin{abstract}
Deep hashing methods have received much attention recently, which achieve promising results by taking advantage of the strong representation power of deep networks. However, most existing deep hashing methods learn a whole set of hashing functions independently, while ignore the correlations between different hashing functions that can promote the retrieval accuracy greatly. Inspired by the sequential decision ability of deep reinforcement learning, we propose a new \textit{Deep Reinforcement Learning approach for Image Hashing (DRLIH)}. Our proposed DRLIH approach models the hashing learning problem as a sequential decision process, which learns each hashing function by correcting the errors imposed by previous ones and promotes retrieval accuracy. To the best of our knowledge, this is the \textit{first} work to address hashing problem from deep reinforcement learning perspective. The main contributions of our proposed DRLIH approach can be summarized as follows: (1) We propose a \textbf{deep reinforcement learning hashing network}. In the proposed network, we utilize recurrent neural network (RNN) as \textit{agents} to model the hashing functions, which take actions of projecting images into binary codes sequentially, so that the current hashing function learning can take previous hashing functions' error into account. (2) We propose a \textbf{sequential learning strategy} based on proposed DRLIH. We define the state as a tuple of internal features of RNN's hidden layers and image features, which can reflect history decisions made by the agents. We also propose an action group method to enhance the correlation of hash functions in the same group. Experiments on three widely-used datasets demonstrate the effectiveness of our proposed DRLIH approach.
\end{abstract}

\begin{IEEEkeywords}
Deep reinforcement learning, image hashing, image retrieval.
\end{IEEEkeywords}

%
\IEEEpeerreviewmaketitle

\section{Introduction}
\IEEEPARstart{W}{ith} rapid growth of images on the web, the large scale image retrieval has attracted much attention. Many hashing methods have been proposed for the fast  image retrieval~\cite{wang2018survey,gionis1999similarity,7298947,wang2010sequential,weiss2009spectral,7937842,chen2017nonlinear,li2013spectral,zhang2014prior,kafai2014discrete,ding2017cross,liong2016deep,hao2017stochastic}. Generally speaking, the goal of hashing methods is to learn several mapping functions, so that the similar images are mapped into similar binary codes. Traditional hashing methods use hand-crafted features (e.g. GIST~\cite{oliva2001modeling}, Bag-of-Visual-Words~\cite{fei2005bayesian}) as image representations, which can not well represent the image content and limit the performance of image retrieval. Inspired by the recent success of deep networks on many computer vision tasks such as image classification and object detection~\cite{krizhevsky2012imagenet}, deep hashing methods~\cite{xia2014supervised,7298947,yaodeep, zhang2015bit,zhao2015deep,li2015feature, zhu2016deep,zhang2017ssdh,zhang2018query} have been proposed to take advantage of feature representation power of deep neural networks.

Existing deep hashing methods~\cite{xia2014supervised,7298947,dsh,hashnet} have achieved promising results on image retrieval. However, they learn the whole set of hashing functions independently, which ignore the correlations between different hashing functions that can promote the retrieval accuracy greatly. There exist several traditional hashing methods~\cite{wang2010sequential,pamisequence} that learn the hash code sequentially, but these methods require complicated optimization that cannot be directly adopted into deep networks.

Recently, deep reinforcement learning has achieved some breakthroughs. For example, deep reinforcement learning achieves human-level performance in Atari games and GO~\cite{Atari,alphago}. A standard reinforcement learning model includes an agent and an environment. The agent receives the information from the environment and chooses the actions to maximize the sum of a reward function. It is noted that the hashing problem has native relationship to the reinforcement learning, which is the motivation of this paper. 1) In the hashing problem, hashing functions can be regarded as agents, which take actions to project images into binary hash codes. 2) The agents in the reinforcement learning framework choose actions to maximize the sum of reward in a task, where the decisions should not be made independently. This property enlightens us that if we regard the hash code generation as a sequential task, it is possible to learn hashing functions dependently within reinforcement learning framework. 

Based on the above analysis, in this paper, we propose a Deep Reinforcement Learning approach for Image Hashing (DRLIH). Instead of learning a whole set of hashing functions independently, our proposed DRLIH approach models the hashing learning process as a sequential decision making process by the designed deep reinforcement learning network. The main contributions of our proposed approach can be summarized as follows:
\begin{itemize}	
	\item \textbf{A Deep reinforcement learning hashing network} is proposed to learn hashing functions sequentially and progressively. The proposed network consists of a feature representation network and a policy network. The policy network is composed by a RNN network, and it serves as the \textit{agent} to sequentially project images into binary codes. We design the policy network to generate the probability of projecting images into hash code $1$, and calculate the probability of hash code $0$. We also propose two \textit{hierarchical reward functions} to drive the training of our proposed DRLIH network.
	\item \textbf{A sequential learning strategy} is proposed to capture the ranking errors caused by previous generated hash codes. We define the states as tuples of the image features and internal features of RNN, which reflect the history decisions, thus the agent can obtain the history information and capture the previous ranking errors to make next decision. We also propose the action group as the minimal step of the agent to enhance the relevance of hash functions in the same group and promote the retrieval accuracy.
\end{itemize}

Experiments on three widely-used datasets demonstrate the effectiveness of the proposed DRLIH approach. The rest of this paper is organized as follows. Section II reviews some representative related works. Section III presents our proposed deep reinforcement learning approach for image hashing, section IV shows the experiments on three widely-used image datasets, and section V concludes this paper.

\section{Related work}
In this section, we briefly review some related works from two aspects: image hashing and deep reinforcement learning.

\subsection{Image Hashing}
The goal of image hashing methods is to project images into Hamming space, where similar images are mapped into similar hash codes to realize efficient image retrieval. Existing hashing methods can be categorized into two classes: data-independent and data-dependent. For the \textit{data-independent methods}, the most representative one is Locality Sensitive Hashing (LSH)~\cite{gionis1999similarity}. LSH uses random projections obtained from Gaussian distributions to map images into binary codes while preserving the cosine similarity. However, LSH needs to generate longer codes and multiple hashing tables to achieve satisfactory performance. Thus some works extend LSH framework to tackle this issue. Lv et al.~\cite{lv2007multi} propose Multi-probe LSH, which uses a multi-probe strategy to avoid generating multiple hashing tables. Raginsky et al.~\cite{raginsky2009locality} propose SIKH to extend LSH into kernel space.

According to the utilization of side information, the \textit{data-dependent methods} can be further classified into unsupervised methods and supervised methods. Unsupervised methods do not require label information to learn hashing functions. For example, Weiss et al.~\cite{weiss2009spectral} propose Spectral Hashing method, which generates balanced hash codes by solving a spectral graph partitioning problem. Liu et al.~\cite{liu2011hashing} propose Anchor Graph Hashing (AGH) method, which learns hashing functions by exploiting the neighborhood structure of data samples by anchor graph. Gong and Lazebnik~\cite{gong2011iterative} present Iterative Quantization (ITQ) method, which simultaneously maximizes the variance of each hash code and minimizes the quantization loss. Zhang et al.~\cite{zhang2013topology} propose Topology Preserving Hashing (TPH) method, which learns hashing functions by preserving not only the neighborhood relationships but also the neighborhood rankings of data points. Irie et al.~\cite{irie2014locally} propose LLH method to model the local linearity of manifolds by locality sensitive sparse coding, which tends to find similar images located in the same manifold as the query. Supervised hashing methods further exploit label information to learn hashing functions for better preserving the semantic similarity of image data. For example, BRE~\cite{kulis2009learning} method is proposed to construct hashing functions to explicitly preserve the original distance (e.g. Euclidean distance) when mapping into Hamming space. Wang et al.~\cite{wang2010sequential} propose SSH method, which learns hashing functions by minimizing the empirical error over labeled data and the information entropy of both labeled and unlabeled data. Liu et al.~\cite{liu2012supervised} propose KSH method, which utilizes the equivalence between optimizing the inner products and the Hamming distances of hash codes, to minimize Hamming distance between similar pairs of data while maximizing the Hamming distance between dissimilar pairs. Shen et al.~\cite{7298598} present SDH method, which expects the generated hash codes to be optimal for classification. Besides pairwise labeled information, some methods also exploit ranking information provided by labels, such as OPH~\cite{wang2013order} and RPH~\cite{wang2015ranking} methods.

The aforementioned learning based methods use hand-crafted features to represent image contents, which limit their retrieval performance. Inspired by the successful application of deep networks in image classification and object detection~\cite{krizhevsky2012imagenet}, some \textit{deep hashing methods}~\cite{xia2014supervised,7298947,yaodeep, zhang2015bit,zhao2015deep,li2015feature, zhu2016deep} are proposed. Xia et al.~\cite{xia2014supervised} propose a two-stage CNNH method, which learns approximate hash codes by preserving the pairwise semantic information in the first hash code learning stage, and then trains a deep hashing network by using the learned hash codes as labels. However, the two-stage scheme causes that the deep networks cannot give feedback for generating better hash codes in the first stage, which limits its retrieval performance. Lai et al.~\cite{7298947} propose NINH method to address this issue. NINH jointly learns the hashing functions and image feature representations simultaneously in Network in Network architecture~\cite{lin2013network}. NINH uses triplet ranking loss~\cite{schultz2003learning,deepranking,jmlrranking} to model the semantic ranking information provided by labels, there are several works follow the one-stage scheme of NINH. For example, BSDH~\cite{zhang2015bit} further learns weights for each hashing function so that the length of hash codes can be determined. Zhu et al.~\cite{zhu2016deep} propose DHN method to further consider the quantization errors caused by hashing layer to promote retrieval accuracy. Yao et al.~\cite{yaodeep} propose DSRH to further consider the orthogonal constraints to make hash codes independent. Zhang et al.~\cite{zhang2017ssdh} propose SSDH method, which trains the deep hashing network in a semi-supervised fashion to enhance the retrieval accuracy and generality of hashing functions. Although deep hashing methods have achieved promising results, they usually learn a whole set of hashing functions independently and directly, which ignore the correlations between different hashing functions that can promote the retrieval accuracy greatly. In this paper, we intend to address this issue from deep reinforcement learning perspective.

\subsection{Deep Reinforcement Learning}
The reinforcement learning is the problem faced by an agent that must learn behavior through trial-and-error interactions with a dynamic environment~\cite{reinforcementsurvey}. In the standard deep reinforcement learning model, the agent receives the current state of the environment as input, and chooses an action as output. The action changes the state of the environment, and the environment communicates to the agent through a scalar reinforcement signal named reward, which reflects the quality of the taken actions. The goal of the reinforcement learning is to train the agent to choose actions that maximize the sum of reward. Recently, deep reinforcement learning methods have achieved some progresses in many domains. Mnih et al.~\cite{Atari} utilize deep neural networks to learn an action-value function to play Atari games, which reaches human-level performance. Silver et al.~\cite{alphago} use policy network and value network to play Go and beat the world-level professional player. Deep reinforcement learning has also been applied in various computer vision tasks. Caicedo et al.~\cite{drlobj} propose a deep reinforcement learning method for active object localization, where the agent is trained to deform the bounding box using sample transformation actions. Zhou et al.~\cite{drlimagecaption} propose a deep reinforcement learning based image caption model, which utilizes a ``policy network" and a ``value network" to collaboratively generate captions. Zhao et al.~\cite{drlcls} utilize the information entropy to guide a reinforcement learning agent to select the key part of an image for better image classification. Rao et al.~\cite{ADRL} propose attention-aware deep reinforcement learning (ADRL) to discard the misleading and confounding frames and focus on the attentions in face videos for better person recognition. Inspired by the recent advances of deep reinforcement learning, we think that hashing problem can be modeled by deep reinforcement learning from two aspects: 1) Hashing functions can be regarded as agents, which take actions to project images into binary hash codes. 2) The agents in the reinforcement learning framework can choose actions to maximize the sum of rewards in a task, which enlightens us that if we regard the hash code generation as a sequential task, it is possible to learn hashing functions dependently within reinforcement learning framework.

\begin{figure*}[ht]
	\centering
	\includegraphics[width=\textwidth]{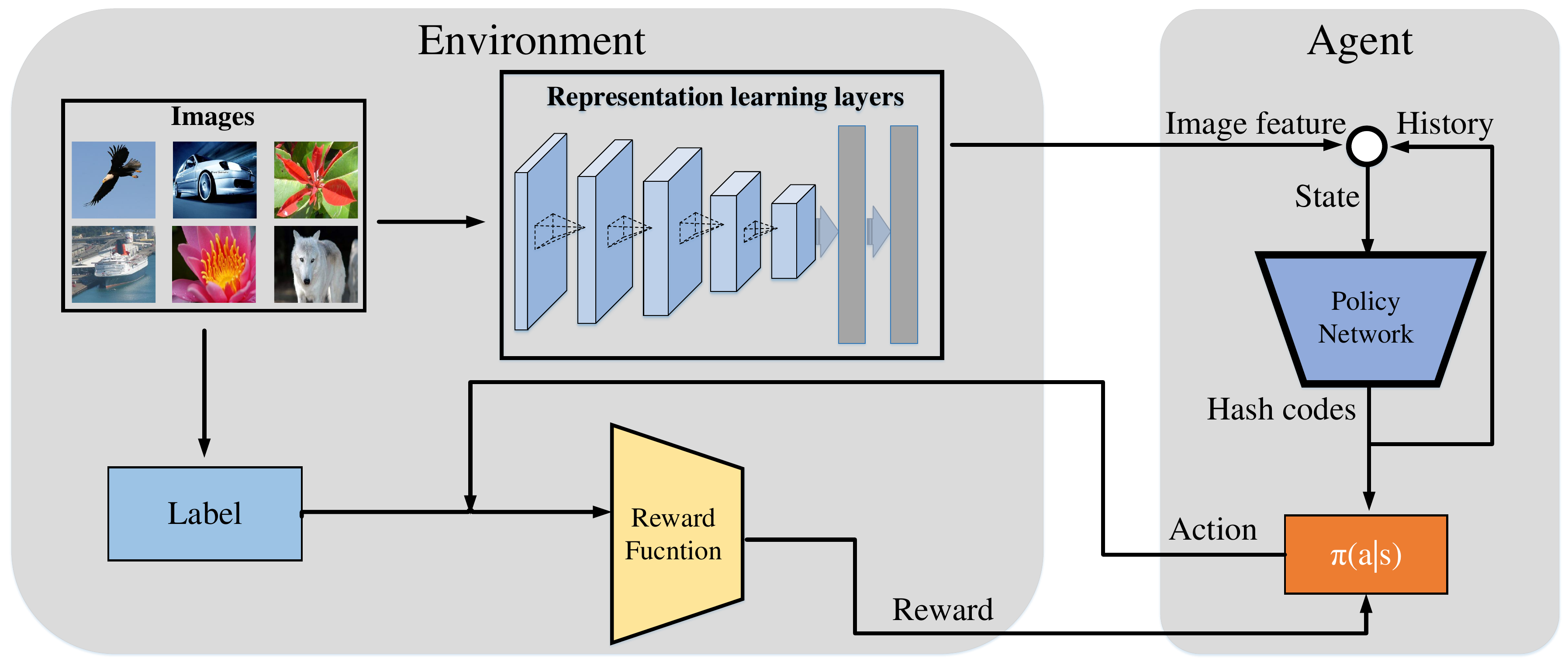}
	\caption{Overview of our proposed deep reinforcement learning hashing network, which consists of a representation learning network, a policy network and a reward function. The policy network serves as the agent and makes decisions to project images into binary codes, while the reward function drives the sequential training of the whole network.}
	\label{framework}
\end{figure*}

\section{Deep reinforcement learning hashing}
Different from existing deep hashing methods, we model the hashing problem as a Markov Decision Process (MDP), which provides a formal framework to model an \textit{agent} that makes a sequence of decisions. In our proposed DRLIH, we consider a batch of images as the \textit{environment}, and define the \textit{states} as image features combined with images' internal features of the policy network. The \textit{agent} projects images into binary codes based on the environment and its current states. In the following part, we first give the formal notations of hashing problems, then we introduce the definition of state, action and reward in our proposed deep reinforcement learning hashing network, then we introduce the network structure of our proposed DRLIH in detail, finally we introduce the sequential learning strategy of DRLIH network.
\subsection{Notations}
Given a set of $n$ images $\mathcal{X} \in \mathbb{R}^D$. The goal of hashing methods is to learn a set of $q$ hashing functions $\mathcal{H}=[h_1,h_2,\cdots,h_q]$, which encode an image $x \in \mathcal{X}$ into a \textit{q}-dimensional binary code $\mathcal{H}(x)$ in the Hamming space, while preserving the semantic similarity of images. Most deep hashing methods learn $\mathcal{H}$ independently and directly, while ignore the correlations between different hashing functions. In this paper we intend to learn the hashing functions $\mathcal{H}$ sequentially by deep reinforcement learning.
\subsection{Definition of the reinforcement hashing learning}
\subsubsection{State} In our DRLIH approach, the current hash code supposes to be generated in serialization which can capture the ranking errors caused by the previous generated hash codes. So the state has to reflect the history information of previous hash codes. We define the state as a tuple $(h,i)$, where $h$ is a history action vector of the generated hash codes, and $i$ denotes the image feature vector. The image feature vector is extracted from the original images using a pre-trained CNN model, and the history action vectors can be obtained from the policy network.
\subsubsection{Action} Given the state tuple $s = (h,i)$, the agent will predict the probability of the actions for current state. It is noted that the hashing problem  only has two possible actions (1 or 0) and the sum of the action probability equals to $1$. Different from most of the reinforcement methods that predict the probability distribution for every possible action, we take the probability of the hash code 1 as the policy network output. The overall probability distribution is formulated as follows:
\begin{equation}
P(a|s,\theta)=\left\{ \begin{array}{rr}
1 - policy(s,\theta) & $a=0$\\
policy(s,\theta) & $a=1$ 
\end{array}
\right.
\end{equation}
where the $policy(s,\theta)$ denotes the output of policy network with input state $s$ and parameters $\theta$.

Considering that only one bit of hash code does not have enough ability to correct the history ranking errors, we propose an action group to address this problem. An action group is composed by k adjacent hash functions. Each action in the group shares the same reward which is designed to enhance the ability of correcting the ranking errors. The action group is the minimal step for the agent which is the actual action definition in the framework. The action probability $\pi_\theta(s_i,A_i)$ is formulated as follows:
\begin{equation}
\begin{split}
& A_i = [a_{t_i+1},a_{t_i+2},..a_{t_i+k}] \\
&\pi_\theta(s_i,A_i) = \prod_{j=1}^{k} P(a_{t_i+j}|\hat{s}_{i,j},\theta)
\end{split}
\end{equation}
where $A_i$ is the $i$-th action group, $a_{t_i+1..k}$ is the element of the action group, $\hat{s}_{i,j}$ is the input state of the $a_{t_i+1}$ and will be given more details in next subsection.
\subsubsection{Reward} After we obtain the generated hash codes, we use a triplet ranking loss~\cite{schultz2003learning,deepranking,jmlrranking} as the reward function to measure the quality of generated codes. It is noted that the probability of the hash code 1 can be regarded as relaxed hash code, which is widely used in many hashing methods~\cite{7298947,zhang2015bit,zhu2016deep}. The higher probability of the action to project images into hash code 1, the value of the probability is closer to 1 and vice versa. So we adopt the probability sequence of projecting images into hash code 1 as the relaxed hash code to calculate the reward.

For training images $(\mathcal{X},\mathcal{Y})$, wehre $\mathcal{Y}$ denotes the corresponding labels. We sample a set of triplet tuples based on the labels, $\mathcal{T}=\{(x_i,x_i^{+},x_i^{-})\}^t_{i=1}$, in which $x_i$ and $x_i^{+}$ are two similar images with the same labels, while $x_i$ and $x_i^{-}$ are two dissimilar images with different labels, and $t$ is the number of sampled triplet tuples. For the triplet tuples $(x_i,x_i^{+},x_i^{-}), i=1 \cdots t$, the reward function is defined as:
\begin{equation}
\label{rankingterm2}
\begin{split}
&\mathcal{J}(h(x_i),h(x_i^{+}),h(x_i^{-})) = \\
&\max(0, m_t+\|h(x_i)-h(x_i^{+})\|^2-\|h(x_i)-h(x_i^{-})\|^2)
\end{split}
\end{equation}
where $\|\cdot\|^2$ denotes the Euclidean distance, $h(\cdot)$ denotes the probability sequence of the corresponding image, and the constant parameter $m_t$ defines the margin between the relative similarities of the two pairs $(x_i,x_i^{+})$ and $(x_i,x_i^{-})$, that is to say, we expect the distance of dissimilar pair $(x_i,x_i^{-})$ to be larger than that of similar pair $(x_i,x_i^{+})$ by a margin of at least $m_t$. Triplet ranking loss based reward can reflect the semantic ranking information, which well evaluates the quality of previous generated hash codes. It is noted that although triplet ranking loss is based on semantic labels, our proposed approach is different from supervised hashing methods. Since we only use triplet ranking loss to evaluate how well the learned hash functions performs on the current environment, then the next hashing function can make decisions to generate hash codes based on the calculated reward, thus our proposed approach fits the reinforcement learning paradigm.

There are two hierarchical rewards to encourage the agent to find the correct hash codes. The first reward is the action group reward which mainly focuses on the hash code quality in the group level. The second reward is the global action reward which focus on the quality of the whole hash code. The hierarchical rewards are defined as:
\begin{equation}
\label{rewards}
\begin{split}
& R^g_{i,j} = -\mathcal{J}(h_j(x_i),h_j(x_i^{+}),h_j(x_i^{-})) \\
& R^G_i = -\mathcal{J}(h(x_i),h(x_i^{+}),h(x_i^{-}))
\end{split}
\end{equation}
where $R^g_{i,j}$ denotes the group action reward of the $i$-th image in $j$-th action group, $h_j$ denotes the probability sequence of the $j$-th action group, and $R^G_i$ denotes the global action reward of the $i-th$ image.

\subsection{Deep reinforcement learning hashing network}
The overall framework of our proposed deep reinforcement learning network is shown in Figure~\ref{framework}, which consists of two parts. The first part is the environment including a representation network and a reward function, which provide the reward and state for the agent. The second part is a policy network serves as the agent, which obtains the state from the environment and generates the hash codes.
\begin{figure*}[htb]
	\centering
	\includegraphics[width=0.9\textwidth]{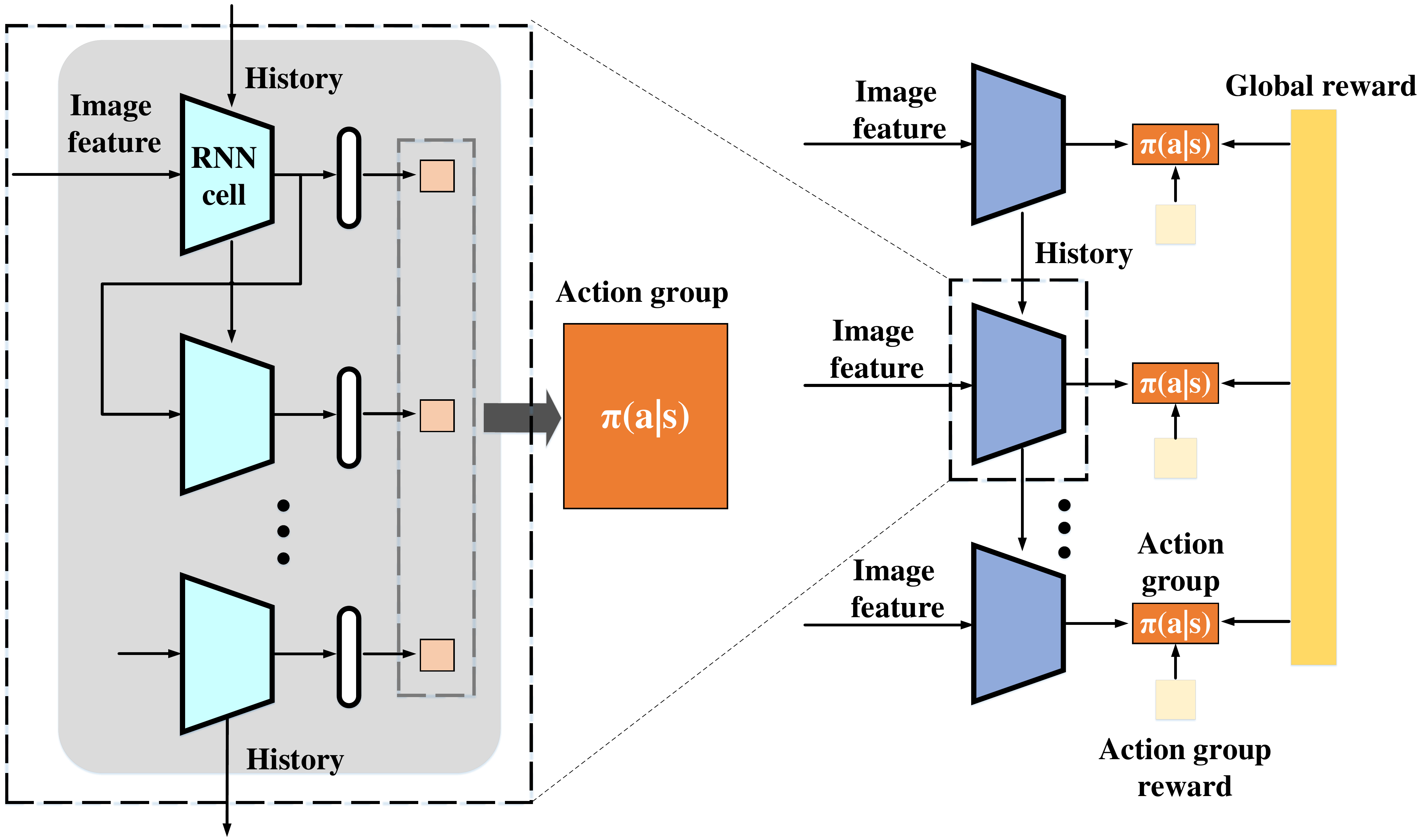}
	\caption{The unrolled details of the policy network. During the training stage, the global reward ensures the generated codes to preserve the semantic ranking information, while the action group reward drives the network to correct the errors caused by previous generated hash codes.}
	\label{sequential}
\end{figure*}
\subsubsection{The representation network} The representation network serves as a feature extractor, it is a deep convolutional network composed of several convolutional layers and fully connected layers. The representation network provides the image features $i$ in the state tuple $(h,i)$. We adopt the VGG-19 network~\cite{Simonyan14c} as the representation network. The first 18 layers follow the exactly same settings in the VGG-19 network. We mainly focus on the design of the policy network and the reward function.
\subsubsection{Policy network}
The policy network is composed of a RNN layer and a policy layer. RNN layer transforms the image features into an internal state, while the policy layer further maps the internal state into a policy probability. The main idea behind the RNN model is a built-in memory cell, which stores information over previous steps and implies the history actions. Figure~\ref{sequential} shows the details of the policy network, which maps the state tuple $(h,i)$ to the probability of an action group. The memory cell aggregates information from two sources: the previous cell memory unit, and the input vector in current step. Formally, for a state tuple $(h,i)$, the RNN layer maps the input to an output sequence by computing activations of the units in the network with the following equation recursively:
\begin{align}
{\bm c}_t &= tanh ({\bm W}_{xi} {\bm x_t} + {\bm b}_{xi} + {\bm W}_{hi} {\bm c}_{t-1} + {\bm b}_{hi})
\end{align}
where ${\bm x_t}, {\bm c}_t$ are the input and hidden vectors, $t$ denotes the $t$-th step, ${\bm W}_{xi}$ and ${\bm W}_{hi}$ are the weight matrix from the input ${\bm x_t}$ and hidden vectors to the new hidden state, ${\bm b}_{xi}$ and ${\bm b}_{hi}$ are the bias terms.

We initialize the RNN layer with $c_0=h$ and $x_0=i$. But in the following step, we set the $x_i=c_{i-1}$ to emphasis the history action information in one action group. Although the image feature is only the input of the first step, the information of the image feature remains in the hidden state with the changing history action information. Thus the hash codes are influenced by the history information and adjusted to correct the errors caused by previous actions. The last hidden state $c_k$ will be regarded as the history information $h$ to synthesis the state of next step.

The policy layer is a fully connect layer defined as:
\begin{equation}
h_t(x) = sigmoid(W_h^Tc_t+v)
\end{equation}
where $c_t$ is the output extracted from RNN layer in step $t$, $W_h$ denotes the weights in the policy layer, and $v$ is the bias parameter. Through the policy layer, the output of RNN at step $t$ is mapped into $[0,1]$. We apply a threshold function to obtain the final binary codes from the policy probability:
\begin{equation}
\label{binarycode}
b_k(x) = g(h(x)) = sgn(h_k(x)-0.5), \quad k=1,\cdots,q
\end{equation}

\subsection{Agent training strategy}
Finally we introduce the agent training strategy. Firstly we unroll the policy network to better show the details, as shown in Figure~\ref{sequential}, we use two reward functions to drive the training of policy network which are defined in equation~(\ref{rewards}).

We use the Monte-Carlo Policy Gradient~\cite{policygradient} to update the parameters to maximize the expected total reward of the action group:
\begin{equation}
\begin{split}
L_g(\theta) &= E_{A\sim\pi(A|s;\theta)}(\sum_{i}R^g_{i}) \\
& \approx \sum_{i}  log[\pi(A_{i}|s_{i};\theta)]R^g_{i} \\
& = \sum_{i} \sum_k  log[P(a_{i,k}|\hat{s}_{i,k};\theta)]  R^g_{i}
\end{split}
\end{equation}
where $L_g(\theta)$ is the expected total reward of the action group.

The global action reward mainly focuses on the quality of the whole set of hash codes, we adopt gradient decent method to optimize the global action reward. According to equation (\ref{rankingterm2}), we can compute the sub-gradient for each triplet tuple $(x_i,x_i^{+},x_i^{-})$, with respect to $h(x_i)$, $h(x_i^{+})$ and $h(x_i^{-})$ as:
\begin{equation}
\begin{split}
&\frac{\partial{R^G_i}}{\partial{h(x_i)}} = 2(h(x_i^{-})-h(x_i^{+}))\times I_{c} \\
&\frac{\partial{R^G_i}}{\partial{h(x_i^{+})}} = 2(h(x_i^{+})-h(x_i))\times I_{c} \\
&\frac{\partial{R^G_i}}{\partial{h(x_i^{-})}} = 2(h(x_i)-h(x_i^{-}))\times I_{c} \\
& I_{c} = I_{m_t+\|h(x_i)-h(x_i^{+})\|^2-\|h(x_i)-h(x_i^{-})\|^2>0}
\end{split}
\end{equation}
where $I_{c}$ is an indicator function, $I_{c} = 1$ if $c$ is true, otherwise $I_{c} = 0$. Thus the global reward can ensure the generated codes to preserve the semantic ranking information.

At last, we explain how the policy network has the ability to correct the history ranking errors through the details of RNN gradient. The gradient of RNN network is formulated as:
\begin{equation}
\begin{split}
&\delta^t_h = \theta'(c_t)(\delta^t_kw_h+\delta^{t+1}_hw_{hi})		\\
&\delta^t_h \overset{\triangle}{=} \frac{\partial\mathcal{J}_L}{\partial c_t} \\
&\delta^t_k \overset{\triangle}{=} \frac{\partial\mathcal{J}_L}{\partial h_t}
\end{split}
\end{equation}
where $c_t$ and $h_t$ are the hidden state and output of RNN respectively, $\theta(c_t)$ denotes the function which maps the $c_t$ to $h_t$. The gradient of hidden layer in the step $t$ consists of two parts: the gradient from the output in $t$ step , and the gradient from the hidden layer in $t+1$ step. The latter is considered as the sequential reward which can correct the errors caused by previous generated hash codes. In the training stage, the previous hash functions will receive the gradient information from current hash function and update the parameters to improve the retrieval accuracy.

\section{Experiments}
In this section, we will introduce the experiments conducted on 3 widely-used datasets compared with 8 state-of-the-art methods, including unsupervised methods LSH~\cite{gionis1999similarity}, SH~\cite{weiss2009spectral}, ITQ~\cite{gong2011iterative} and supervised methods SDH~\cite{7298598}, CNNH~\cite{xia2014supervised}, NINH~\cite{7298947}, DSH~\cite{dsh} and HashNet~\cite{hashnet}. LSH, SH, SDH, and ITQ are traditional hashing methods without deep networks, while CNNH, NINH, DSH, HashNet and our proposed DRLIH are deep hashing methods, which take the raw image pixels as input to conduct hashing function learning.

\subsection{Datasets}
We conduct experiments on 3 widely-used image retrieval datasets:
\begin{itemize}
	\item The \textbf{CIFAR10}~\cite{cifar10} dataset contains 60000 color images from 10 classes, the size of each image is $32\times 32$. Following~\cite{xia2014supervised,7298947}, we randomly select 1000 images as query set. For the compared unsupervised methods, all the rest images are used as the training set, while for the compared supervised methods, we further randomly select 5000 images to form the training set.
	\item The \textbf{NUS-WIDE}~\cite{chua2009nus} dataset contains nearly 270000 images with associated labels from 81 semantic concepts. We use the 21 most frequent concepts to conduct experiment following~\cite{7298947}. We randomly select 2100 images as the query set, 100 images per concept. All the rest images are used as training set for unsupervised methods, while we further randomly select 10500 images to form training set for supervised methods.
	\item The \textbf{MIRFlickr}~\cite{huiskes2008mir} dataset consists of 25000 images obtained from Flickr website as well as associated tags. These images are annotated with one or multiple labels of 38 semantic concepts. Similarly, we randomly choose 1000 images as the query set, and use the rest images as the training set of unsupervised methods, and further select 5000 images as training set of supervised methods.
\end{itemize}
\begin{table*}[htb]
	\centering
	\caption{MAP scores with different length of hash codes on CIFAR10, NUS-WIDE and MIRFlickr datasets. MAP scores are calculated based on top 5000 returned images for NUS-WIDE dataset. The best results of each code length are shown in boldface.}
	\label{maptable}
	\begin{tabularx}{\textwidth}{c|YYYY|YYYY|YYYY}
		\hline
		\multirow{2}{*}{Methods} & \multicolumn{4}{c|}{CIFAR10}  & \multicolumn{4}{c|}{NUS-WIDE}  & \multicolumn{4}{c}{MIRFlickr} \\ \cline{2-13} 
		& 12bit & 24bit & 32bit & 48bit & 12bit   & 24bit   & 32bit   & 48bit & 12bit   & 24bit   & 32bit   & 48bit  \\ \hline
		\textbf{DRLIH (ours)}             		& \textbf{0.816} & \textbf{0.843} & \textbf{0.855} & \textbf{0.853} & \textbf{0.823}   & \textbf{0.846}   & \textbf{0.845}   & \textbf{0.853} & \textbf{0.796}   & \textbf{0.811}   & \textbf{0.810}   & \textbf{0.814} \\ 
		HashNet$\ast$			& 0.765 & 0.823 & 0.840 & 0.843 & 0.812   & 0.833   & 0.830   & 0.840  & 0.777  & 0.782 & 0.785 & 0.785\\ 
		DSH$\ast$      			    & 0.708 & 0.712 & 0.751 & 0.720  & 0.793   & 0.804   & 0.815 & 0.800  & 0.651	& 0.681	& 0.684	& 0.686
		\\
		NINH$\ast$			      	& 0.792 & 0.818 & 0.832 & 0.830 & 0.808   & 0.827   & 0.827   & 0.827 & 0.772 	& 0.756 &	0.760 &	0.778 
		\\ 
		CNNH$\ast$			      	& 0.683 & 0.692 & 0.667 & 0.623 & 0.768   & 0.784   & 0.790   & 0.740 & 0.763 	& 0.757 &	0.758 &	0.755 
		\\ \hline
		SDH-VGG19            		& 0.430 & 0.652 & 0.653 & 0.665 & 0.730   & 0.797   & 0.819   & 0.830 & 0.732 	& 0.739 &	0.737 &	0.747 
		\\ 
		ITQ-VGG19            		& 0.339 & 0.361 & 0.368 & 0.375 & 0.777   & 0.800   & 0.806   & 0.817 & 0.686 	& 0.685 &	0.687 &	0.689 
		\\ 
		SH-VGG19             		& 0.244 & 0.213 & 0.213 & 0.209 & 0.712   & 0.697   & 0.689   & 0.682 & 0.618 	& 0.604 &	0.598 &	0.595 
		\\ 
		LSH-VGG19            		& 0.133 & 0.171 & 0.178 & 0.198 & 0.518   & 0.567   & 0.618   & 0.651 & 0.575 	& 0.584 &	0.604 & 0.614 
		\\ \hline
		SDH        					& 0.255 & 0.330 & 0.344 & 0.360 & 0.460   & 0.510   & 0.519   & 0.525  & 0.595  & 0.601 & 0.608 & 0.605\\ 
		ITQ     					& 0.158 & 0.163 & 0.168 & 0.169 & 0.472   & 0.478   & 0.483   & 0.476  & 0.576  & 0.579 & 0.579 & 0.580\\ 
		SH         					& 0.124 & 0.125 & 0.125 & 0.126 & 0.452   & 0.445   & 0.443   & 0.437  & 0.561  & 0.562 & 0.563 & 0.562\\ 
		LSH        					& 0.116 & 0.121 & 0.124 & 0.131 & 0.436   & 0.414   & 0.432   & 0.442  & 0.557  & 0.564 & 0.562 & 0.569\\ \hline
	\end{tabularx}
\end{table*}
\begin{figure*}[tbh]
	\centering
	\includegraphics[width=\textwidth]{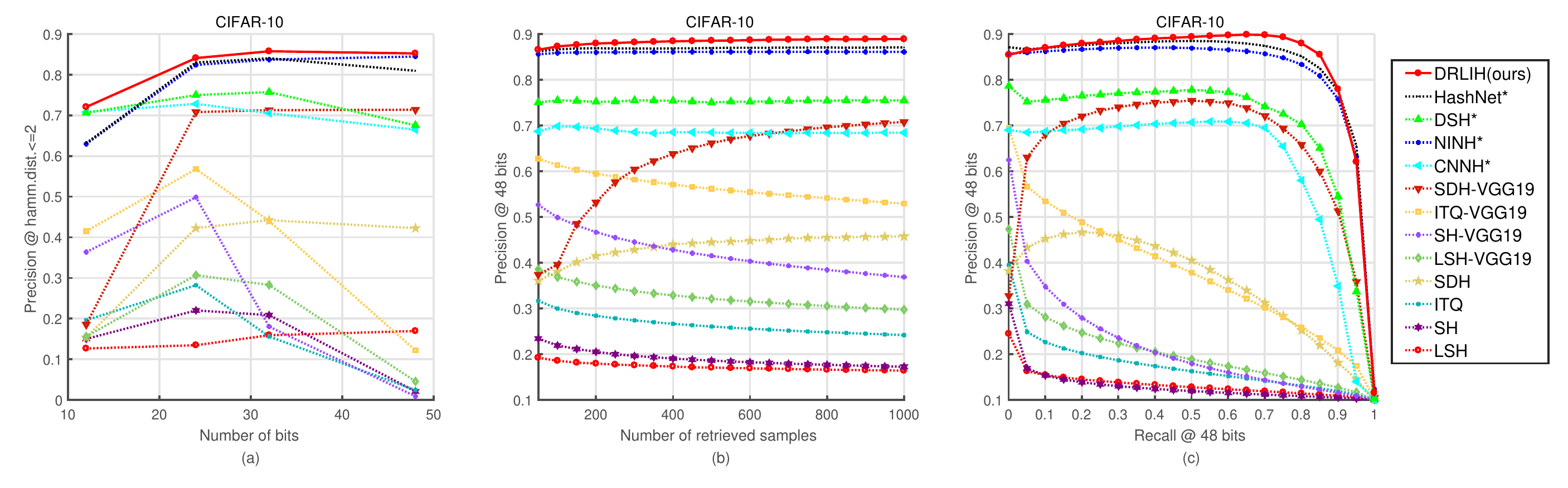}
	\caption{The comparison results on \textbf{CIFAR10}. (a) Precision within Hamming radius 2 using hash lookup; (b) Precision at top $k$ returned results. (c) Precision-Recall curves of Hamming Ranking with 48bit.} 
	\label{CIFAR10Result}
\end{figure*}
\begin{figure*}[tbh]
	\centering
	\includegraphics[width=\textwidth]{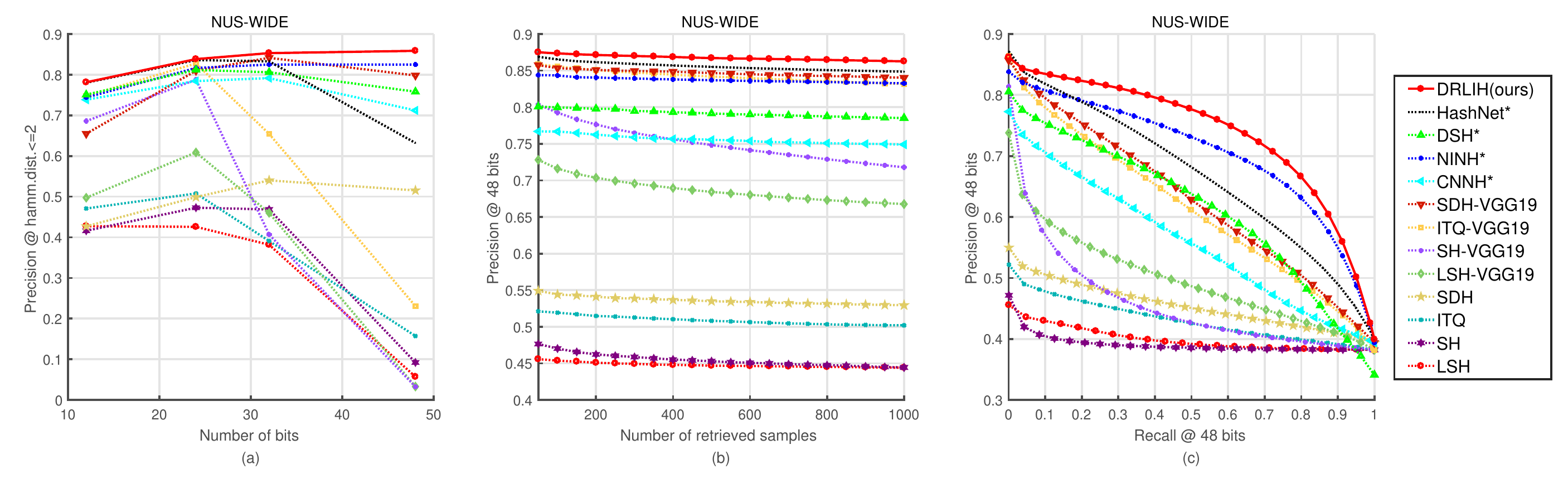}
	\caption{The comparison results on \textbf{NUS-WIDE}. (a) Precision within Hamming radius 2 using hash lookup; (b) Precision at top $k$ returned results. (c) Precision-Recall curves of Hamming Ranking with 48bit.}
	\label{NUSWIDEResults}
\end{figure*}
\begin{figure*}[tbh]
	\centering
	\includegraphics[width=\textwidth]{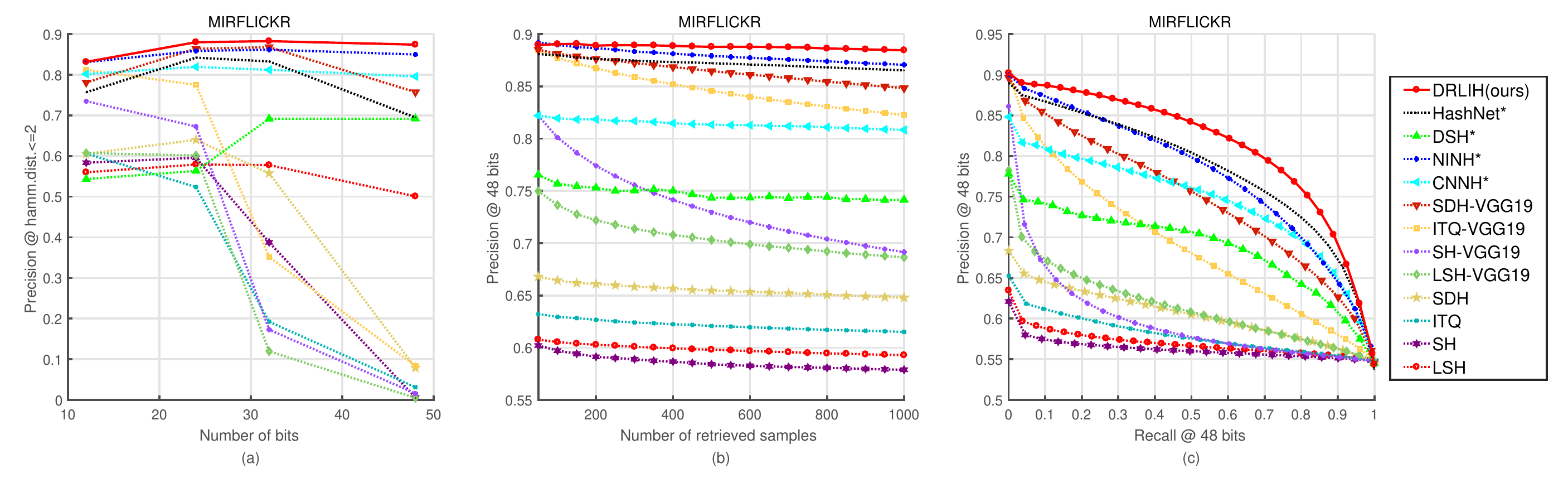}
	\caption{The comparison results on \textbf{MIRFlickr}. (a) Precision within Hamming radius 2 using hash lookup; (b) Precision at top $k$ returned results. (c) Precision-Recall curves of Hamming Ranking with 48bit.}
	\label{MIRFLICKR}
\end{figure*}
\begin{figure*}[htb]
	\centering
	\includegraphics[width=0.9\textwidth]{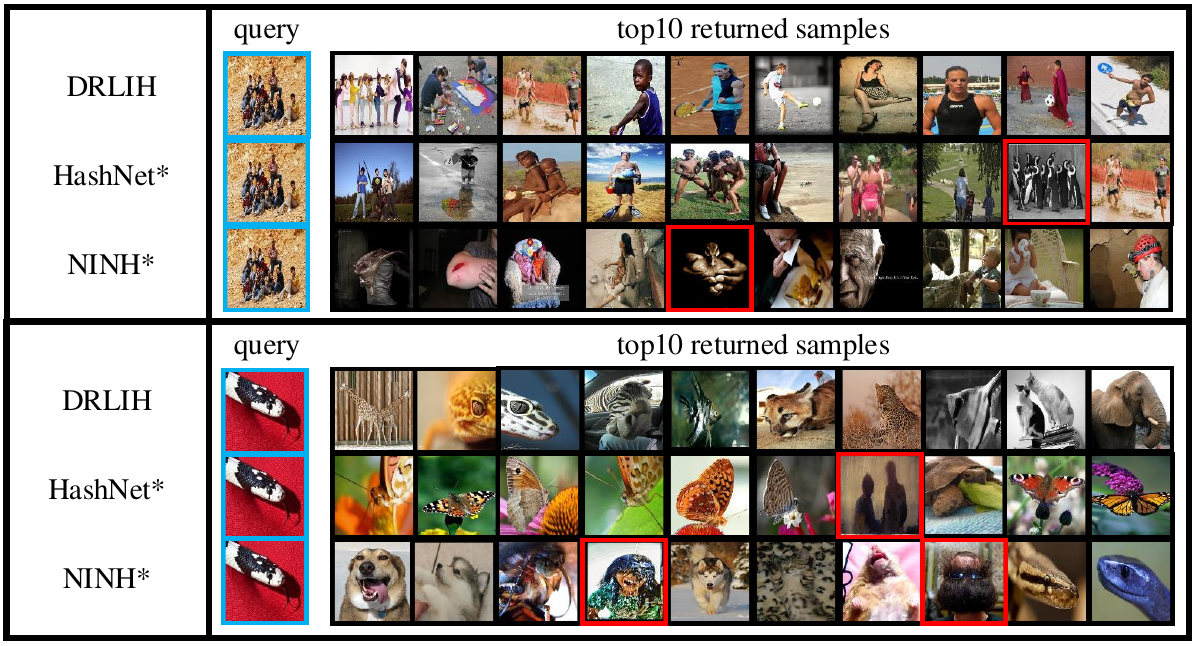}
	\caption{Some retrieval results of NUS-WIDE dataset using Hamming ranking on 48bit hash codes. The blue rectangles denote the query images. The red rectangles indicate wrong retrieval results. We can observe that DRLIH achieves the best results.}
	\label{demofig}
\end{figure*}
\subsection{Experiment Settings and Evaluation Metrics}
We implement the proposed approach based on the open-source framework pytorch\footnote{http://pytorch.org}. The parameters of the first 18 layers in our network are initialized with the VGG-19 network~\cite{Simonyan14c}, which is pre-trained on the ImageNet dataset~\cite{ILSVRC15}. Similar initialization strategy has been used in other deep hashing methods~\cite{zhu2016deep,yaodeep,dsh}. The dimension of RNN's hidden layer is set to be 4096 in the policy network. In all experiments, our networks are trained with the initial learning rate of 0.001, we decrease the learning rate by a factor of 10 every 10000 steps. And the mini-batch size is 16, the weight decay parameter is 0.0005. For the parameter in our proposed loss function, we set $m_t=1$ in all the experiments. For the length of action group, we set it as $12$ through out the experiments.

We compare the proposed DRLIH approach with eight state-of-the-art methods, including unsupervised methods LSH, SH and ITQ, supervised methods SDH, CNNH, NINH, DSH and HashNet. The brief introductions of these methods are as follows: 
\begin{itemize}
	\item \textbf{LSH}~\cite{gionis1999similarity} is a data independent unsupervised method, which uses randomly generated hash functions to map image features into binary codes.
	\item \textbf{SH}~\cite{weiss2009spectral} is a data dependent unsupervised method, which learns hash functions by making hash codes balanced and uncorrelated.
	\item \textbf{ITQ}~\cite{gong2011iterative} is also a data dependent unsupervised method, which learns hash functions by minimizing the quantization error of mapping data to the vertices of a binary hypercube.
	\item \textbf{SDH}~\cite{7298598} is a supervised method, which leverages label information to obtain hash codes by integrating hash code generation and classifier training.
	\item \textbf{CNNH}~\cite{xia2014supervised} is a two-stage deep hashing method, which learns hash codes for training images in first stage, and trains a deep hashing network in second stage.
	\item \textbf{NINH}~\cite{7298947} is a one-stage deep hashing method, which learns deep hashing network by a triplet loss function to measure the ranking information provided by labels.
	\item \textbf{DSH}~\cite{dsh} is a one-stage deep hashing method, which designs a loss function to maximize the discriminability of the output space by encoding the supervised information from the input image pairs, and simultaneously imposing regularization on the real-valued outputs to approximate the desired discrete values.
	\item \textbf{HashNet}~\cite{hashnet} directly learns the binary hash codes and addresses the ill-posed gradient and data imbalance problems in an end-to-end framework of deep feature learning and binary hash encoding.
\end{itemize}

For our proposed DRLIH, and compared CNNH, NINH, DSH and HashNet methods, we use the raw image pixels as input. The implementations of CNNH, DSH and HashNet are provided by their authors, while NINH is our own implementation. Since the representation learning layers of CNNH, NINH, DSH and HashNet are different from each other, for a fair comparison, we use the same VGG-19 network as the base structure for all the deep hashing methods. And the network parameters of all the deep hashing methods are initialized with the same pre-trained VGG-19 model, thus we can perform a fair comparison between them. The results of CNNH, NINH, DSH and HashNet are referred as CNNH$\ast$, DSH$\ast$, NINH$\ast$ and HashNet$\ast$ respectively.

For other compared traditional methods without deep networks, we represent each image by hand-crafted features and deep features respectively. For hand-crafted features, we represent images in the CIFAR10 and MIRFlickr by 512-dimensional GIST features, and images in the NUS-WIDE by 500-dimensional bag-of-words features. For a fair comparison between traditional methods and deep hashing methods, we also conduct experiments on the traditional methods with deep features, where we extract 4096-dimensional deep feature for each image from the same pre-trained VGG-19 network. We denote the results of traditional methods using deep features by LSH-VGG19, SH-VGG19, ITQ-VGG19 and SDH-VGG19. The results of SDH, SH and ITQ are obtained from the implementations provided by their authors, while the results of LSH are from our own implementation.

To objectively and comprehensively evaluate the retrieval accuracy of the proposed approach and the compared methods, we use four evaluation metrics: Mean Average Precision (MAP), precision at top $k$ returned results, precision-recall curves and precision within Hamming radius 2 using hash lookup. These four evaluation metrics are defined as follows:
\begin{itemize}
	\item The MAP scores are computed as the mean of average precision (AP) for all queries, and AP is computed as:
	\begin{equation}
	AP=\frac{1}{R}\sum_{k=1}^{n}\frac{k}{R_k}\times rel_k
	\end{equation}
	where $n$ is the size of database, R is the number of relevant images in database, $R_k$ is the number of relevant images in the top $k$ returns, and $rel_k=1$ if the image ranked at $k$-th position is relevant and 0 otherwise.
	\item Precision at top $k$ returned results (top$K$-precision): The precision with respect to different numbers of retrieved samples from the ranking list.
	\item Precision within Hamming radius 2: Precision curve of returned images with the Hamming distance smaller than 2 using hash lookup.
\end{itemize}

\subsection{Experiment Results}
\subsubsection{Experiment results on CIFAR10 dataset}
Table~\ref{maptable} shows the MAP scores with different length of hash codes on CIFAR10 dataset. Overall, the proposed DRLIH achieves the highest average MAP of 0.842, and consistently outperforms state-of-the-art methods on all hash code lengths. More specifically, the result tables are partitioned into three groups: deep hashing methods, traditional methods with deep features and traditional methods with hand-crafted features. Compare with the highest deep hashing methods HashNet$\ast$, which achieves average MAP of 0.818, the proposed DRLIH has an absolute improvement of 0.024. Compare with the highest traditional methods using deep features SDH-VGG19, which achieves an average MAP of 0.600, the proposed method has an absolute improvement of 0.242. While the highest traditional methods using handcrafted features SDH achieves average MAP of 0.322, the proposed approach has an absolute improvement of 0.520.

Figure~\ref{CIFAR10Result}(a) shows the precisions within Hamming radius 2 using hash lookup. The precision of proposed DRLIH consistently outperforms state-of-the-art methods on all hash code lengths. The precision of most traditional methods decrease when using longer hash codes. This is because the number of images sharing the same Hamming code decreases exponentially for longer hash codes (e.g. 48bit), which will cause some queries fail to return images within Hamming radius 2. While the proposed DRLIH achieves the highest precision on 48bit code length, which shows the robustness of proposed method on longer hash codes. Figure~\ref{CIFAR10Result}(b) shows the precision at top $k$ returned results, we can also observe that our proposed DRLIH achieves the best precision compared with state-of-the-art methods. Figure~\ref{CIFAR10Result}(c) demonstrates the precision-recall curves using Hamming ranking with 48bit codes. DRLIH still achieves the best accuracy on all recall levels, which further demonstrates the effectiveness of the proposed approach.
\begin{figure*}[tbh]
	\centering
	\includegraphics[width=\textwidth]{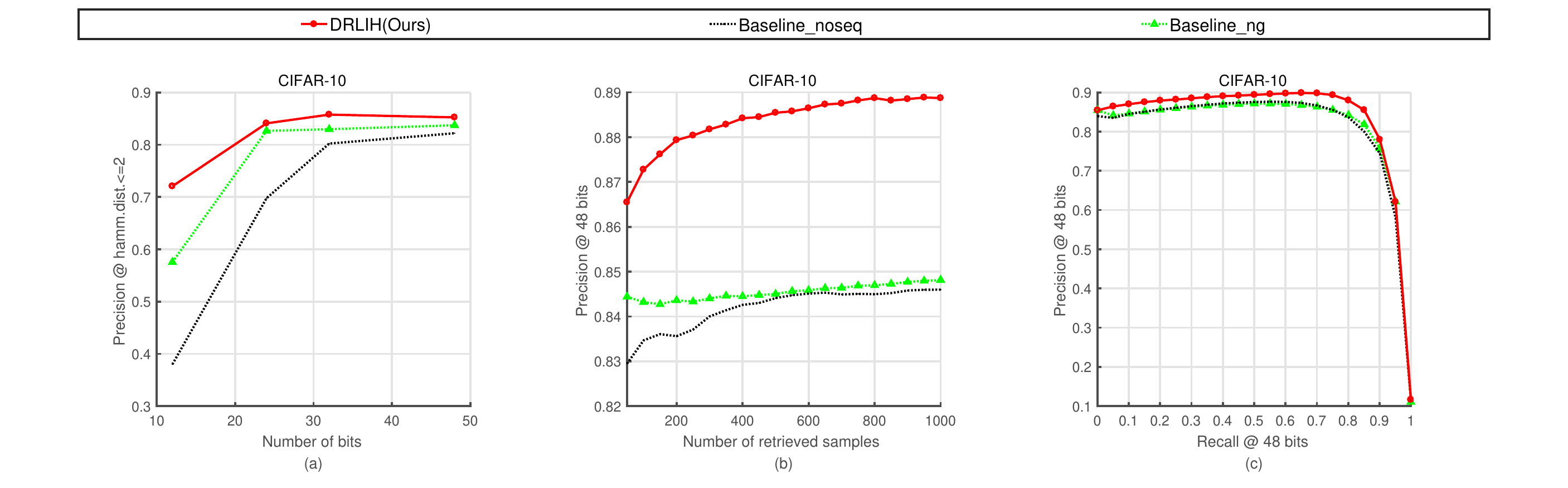}
	\caption{The comparison results of baseline methods on \textbf{CIFAR10}. (a) Precision within Hamming radius 2 using hash lookup; (b) Precision at top $k$ returned results; (c) Precision-Recall curves of Hamming Ranking with 48bit.}
	\label{ciar10baseline}
\end{figure*}
\begin{figure*}[tbh]
	\centering
	\includegraphics[width=\textwidth]{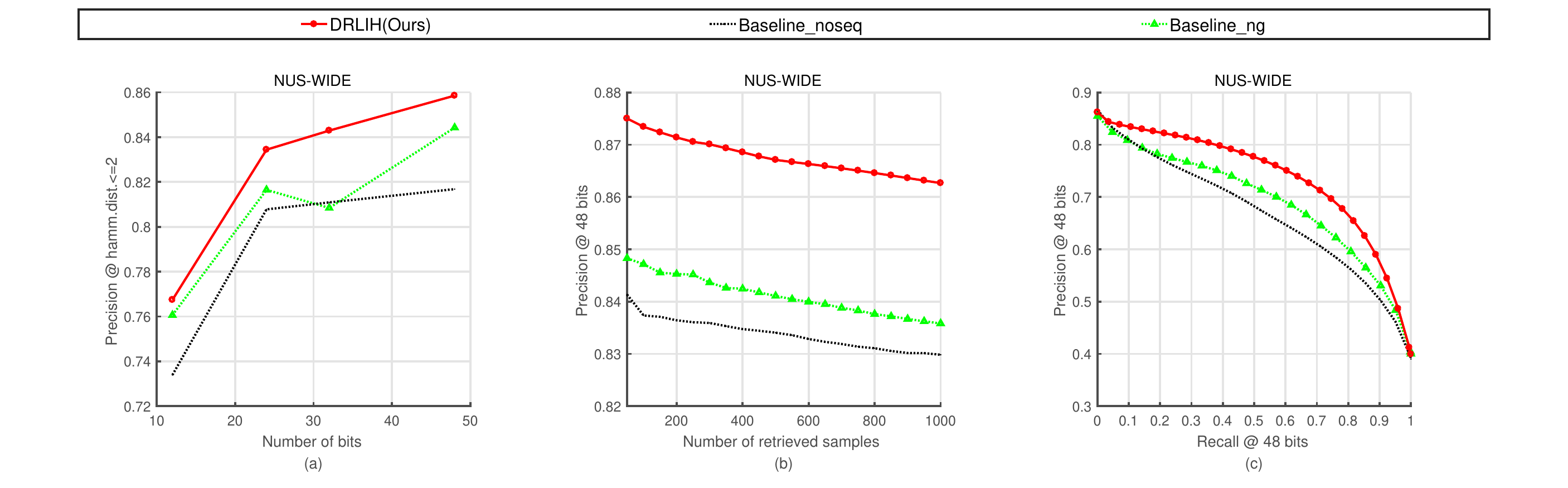}
	\caption{The comparison results of baseline methods on \textbf{NUS-WIDE}. (a) Precision within Hamming radius 2 using hash lookup; (b) Precision at top $k$ returned results; (c) Precision-Recall curves of Hamming Ranking with 48bit.}
	\label{nuswidebaseline}
\end{figure*}
\begin{figure*}[tbh]
	\centering
	\includegraphics[width=\textwidth]{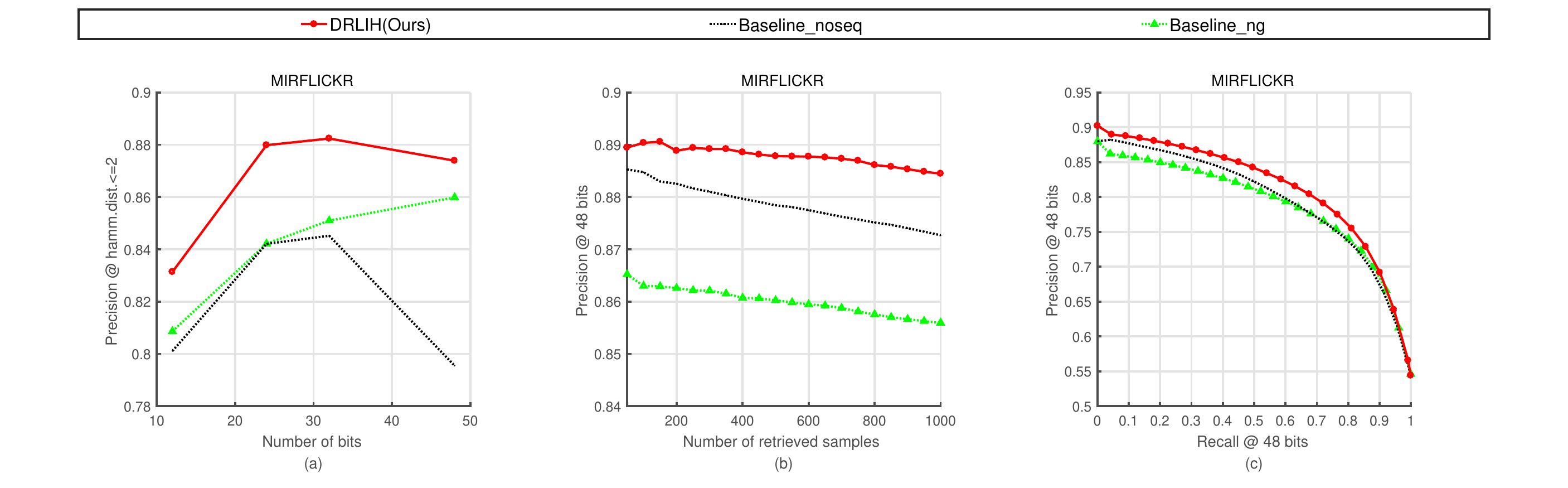}
	\caption{The comparison results of baseline methods on \textbf{MIRFlickr}. (a) Precision within Hamming radius 2 using hash lookup; (b) Precision at top $k$ returned results; (c) Precision-Recall curves of Hamming Ranking with 48bit.}
	\label{flickr25kbaseline}
\end{figure*}
\begin{table*}[htb]
	\centering
	\caption{Comparison between baseline methods and our proposed approach with different length of hash codes on CIFAR10, NUS-WIDE and MIRFlickr datasets.}
	\label{baseline}
	\begin{tabularx}{\textwidth}{c|YYYY|YYYY|YYYY}
		\hline
		\multirow{2}{*}{Methods} & \multicolumn{4}{c|}{CIFAR10}  & \multicolumn{4}{c}{NUS-WIDE}  & \multicolumn{4}{c}{MirFlickr} \\ \cline{2-13} 
		& 12bit & 24bit & 32bit & 48bit & 12bit   & 24bit   & 32bit   & 48bit & 12bit   & 24bit   & 32bit   & 48bit  \\ \hline
		DRLIH (ours)         	& \textbf{0.816} & \textbf{0.843} & \textbf{0.855} & \textbf{0.853} & \textbf{0.823}   & \textbf{0.846}   & \textbf{0.845}   & \textbf{0.853}  & \textbf{0.796}   & \textbf{0.811}   & \textbf{0.810}   & \textbf{0.814} \\ 
		Baseline\_noseq		& 0.759 & 0.802 & 0.811 & 0.822      	& 0.803 & 0.811 & 0.820 & 0.824 & 0.764   & 0.775   & 0.776   & 0.784  \\ 
		Baseline\_ng	    & 0.805 & 0.823 & 0.820 & 0.826      	& 0.811 & 0.824 & 0.829 & 0.835 & 0.774   & 0.778   & 0.784   & 0.789  \\		\hline
	\end{tabularx}
\end{table*}
\subsubsection{Experiment results on NUS-WIDE dataset}
Table~\ref{maptable} shows the MAP scores with different length of hash codes on NUS-WIDE dataset. Following~\cite{xia2014supervised}, we calculate the MAP scores based on top 5000 returned images. Similar results can be observed, our proposed DRLIH still achieves the best MAP scores (average 0.842). DRLIH achieves an absolute improvement of 0.013 on average MAP compared to the highest deep methods HashNet$\ast$ (average 0.829). Compare with the highest traditional method using deep features ITQ-VGG19, which achieves an average MAP of 0.800, DRLIH has an absolute improvement of 0.042. It is also interesting to observe that with the deep features extracted from VGG-19 network, the traditional methods such as SDH and ITQ achieve comparable results with deep hashing methods.

Figures~\ref{NUSWIDEResults}(a), (b) and (c) demonstrate the precision within Hamming radius 2 using hash lookup, precision at top $k$ returned results and the precision-recall curves using Hamming ranking with 48 bits. Similar trends can be observed on these three evaluation metrics, the proposed DRLIH also achieves promising results on NUS-WIDE dataset on all hash code lengths, which further shows the effectiveness of our proposed DRLIH.

\subsubsection{Experiment results on MIRFLICKR dataset}
Table~\ref{maptable} shows the MAP scores with different length of hash codes on MIRFlickr dataset. We can also observe that our proposed DRLIH still achieves the best MAP scores (average 0.808). DRLIH achieves an absolute improvement of 0.026 on average MAP compared to the highest deep methods HashNet$\ast$ (average 0.782). Compare with the highest traditional method using deep features SDH-VGG19, which achieves an average MAP of 0.739, DRLIH has an absolute improvement of 0.069. Compare with the highest traditional method using hand-crafted features SDH, which achieves an average MAP of 0.602, DRLIH has an absolute improvement of 0.206.

We can observe from Figure~\ref{NUSWIDEResults}(a) that the proposed DRLIH achieves the best precision within Hamming radius 2 using hash lookup. Figure~\ref{NUSWIDEResults}(b) demonstrates the precision at top $k$ returned results, our proposed DRLIH achieves the best precision compared with state-of-the-art methods. Figure~\ref{NUSWIDEResults}(c) shows the precision-recall curves using Hamming ranking with 48 bits, and similar trends can be observed that the proposed DRLIH also achieves best precision on all recall levels compared with state-of-the-art methods.

Finally, we demonstrate the top 10 retrieval results of NUS-WIDE using Hamming ranking on 48bit hash codes. As shown in Figure~\ref{demofig}, our proposed DRLIH achieves the best results.

\subsubsection{Baseline experiments}
To verify the effectiveness of action group and sequence learning strategy of deep reinforcement learning, we also conduct two baseline experiments. More specifically, we first set the length of the action group as 1, which implies that we train the network without action group, we denote this method as Baseline\_ng. We also conduct a baseline experiment without the sequence learning strategy, where we replace the agent with a fully-connected layer whose dimension is the same as the hash code length and train the network by the triplet ranking loss. This layer serves as hashing layer that maps features of representation network into binary codes directly and independently. We denote this method as Baseline\_noseq. Comparing the proposed DRLIH approach with Baseline\_ng, we can verify the effectiveness of action group. Comparing our DRLIH approach with Baseline\_noseq, we can verity the effectiveness of the sequence learning strategy. The results are shown in table~\ref{baseline}, and we can observe that compare with Baseline\_ng, our proposed DRLIH approach improves the average MAP score from 0.819 to 0.842 on CIFAR10 dataset, from 0.825 to 0.842 on NUS-WIDE dataset and from 0.781 to 0.808 on MIRFlickr dataset. This demonstrates that the proposed action group method can promote the retrieval accuracy. Comparing with Baseline\_noseq, our proposed DRLIH approach improves the average MAP score from 0.799 to 0.842 on CIFAR10 dataset, from 0.815 to 0.842 on NUS-WIDE dataset and from 0.775 to 0.808 on MIRFlickr dataset. This demonstrates that we can benefit from sequence learning of deep reinforcement learning framework to promote retrieval accuracy. We can also observer that Baseline\_ng has better performance than the Baseline\_noseq, which further shows the effectiveness of the sequential learning strategy. Figures~\ref{ciar10baseline},~\ref{nuswidebaseline} and~\ref{flickr25kbaseline} show the precision within hamming radius 2, precision at top $k$ returned results and precision recall curves on three datasets, we can observe that the proposed DRLIH approach outperforms two baseline methods on these three evaluation metrics.

\section{Conclusion}
In this paper, we have proposed a \textit{Deep Reinforcement Learning approach for Image Hashing (DRLIH)}. First, we propose a policy based deep reinforcement learning network for modeling hashing functions. We utilize recurrent neural network (RNN) to model hashing functions as agents, which take actions of projecting images into binary codes sequentially. While we regard hash codes and the image features as states, which provide history actions taken by agents. The whole network is trained by optimizing two hierarchical reward functions. Second, we propose a sequential learning strategy based on proposed DRLIH, which can iteratively optimize the overall accuracy by correcting the error generated by history actions. Experiments on three widely used datasets demonstrate the effectiveness of our proposed approach.

The future work lies in two aspects: First, we will try to define the sequential learning process explicitly, such that we can better model the sequential learning. Second, we intend to exploit more advanced deep reinforcement learning framework to achieve better retrieval accuracy.


%





\ifCLASSOPTIONcaptionsoff
  \newpage
\fi



%
\bibliographystyle{IEEEtran}
\bibliography{drlih}

%








\end{document}